\documentclass[acmtog]{acmart}

\usepackage[ruled,vlined]{algorithm2e}
\usepackage{multirow}
\usepackage{hhline}
\usepackage{graphicx}
\usepackage{pifont}
\usepackage{verbatim}
\usepackage{booktabs}
\usepackage{esvect}
\usepackage{makecell}
\usepackage{hyperref}
\usepackage{siunitx}
\newcommand{\tabincell}[2]{\begin{tabular}{@{}#1@{}}#2\end{tabular}}
\usepackage{colortbl}

\usepackage{xcolor}
\usepackage{tabularray}

\DeclareSIUnit[quantity-product = ]\percent{\char`\%}

\AtBeginDocument{%
  \providecommand\BibTeX{{%
    \normalfont B\kern-0.5em{\scshape i\kern-0.25em b}\kern-0.8em\TeX}}}

\acmConference[SIGGRPAH ASIA'24]{SIGGRPAH ASIA'24}{December 3--6, 2024}{Tokyo}

\copyrightyear{2024}
\acmYear{2024}
\setcopyright{acmlicensed}\acmConference[SA Conference Papers '24]{SIGGRAPH Asia 2024 Conference Papers}{December 3--6, 2024}{Tokyo, Japan}
\acmBooktitle{SIGGRAPH Asia 2024 Conference Papers (SA Conference Papers '24), December 3--6, 2024, Tokyo, Japan}
\acmDOI{10.1145/3680528.3687615}
\acmISBN{979-8-4007-1131-2/24/12}

\definecolor{orchid}{rgb}{0.85, 0.44, 0.84}

\newcommand{\CHANGE}[1]{{{#1}}}

\begin{document}

\title{RoMo: A Robust Solver for Full-body Unlabeled Optical Motion Capture}

\author{Xiaoyu Pan}
\email{panxiaoyu6@gmail.com}

\author{Bowen Zheng}
\email{zhengbowen.crist@gmail.com}

\affiliation{%
  \institution{State Key Lab of CAD \& CG, Zhejiang University; ZJU-Tencent Game and Intelligent Graphics Innovation Technology Joint Lab}
  \city{Hangzhou}
  \country{China}}

\author{Xinwei Jiang}
\email{wesleyjiang@tencent.com}

\author{Zijiao Zeng}
\email{jasonglxu@tencent.com}

\affiliation{%
  \institution{Tencent Games Digital Content Technology Center}
  \city{Shanghai}
  \country{China}}

\author{Qilong Kou}
\email{rambokou@tencent.com}

\affiliation{%
  \institution{ Tencent Technology (Shenzhen) Co., LTD}
  \city{Shenzhen}
  \country{China}}

\author{He Wang}
\email{he_wang@ucl.ac.uk}

\affiliation{%
  \institution{Department of Computer Science and UCL Centre for Artificial Intelligence, University College London}
  \city{London}
  \country{United Kingdom}}

\author{Xiaogang Jin}
\authornote{Corresponding author.}
\affiliation{%
  \institution{State Key Lab of CAD \& CG, Zhejiang University; ZJU-Tencent Game and Intelligent Graphics Innovation Technology Joint Lab}
  \city{Hangzhou}
  \country{China}}
\email{jin@cad.zju.edu.cn}

\begin{abstract}

Optical motion capture (MoCap) is the "gold standard" for accurately capturing full-body motions. To make use of raw MoCap point data, the system \textbf{labels} the points with corresponding body part locations and \textbf{solves} the full-body motions. However, MoCap data often contains mislabeling, occlusion and positional errors, requiring extensive manual correction. To alleviate this burden, we introduce RoMo, a learning-based framework for robustly labeling and solving raw optical motion capture data. In the labeling stage, RoMo employs a divide-and-conquer strategy to break down the complex full-body labeling challenge into manageable subtasks: alignment, full-body segmentation and part-specific labeling. To utilize the temporal continuity of markers, RoMo generates marker tracklets using a K-partite graph-based clustering algorithm, where markers serve as nodes, and edges are formed based on positional and feature similarities. For motion solving, to prevent error accumulation along the kinematic chain, we introduce a hybrid inverse kinematic solver that utilizes joint positions as intermediate representations and adjusts the template skeleton to match estimated joint positions. We demonstrate that RoMo achieves high labeling and solving accuracy across multiple metrics and various datasets. Extensive comparisons show that our method outperforms state-of-the-art research methods. On a real dataset, RoMo improves the F1 score of hand labeling from 0.94 to 0.98, and reduces joint position error of body motion solving by 25\%. Furthermore, RoMo can be applied in scenarios where commercial systems are inadequate. The code and data for RoMo are available at \textcolor{magenta}{\textit{\url{https://github.com/non-void/RoMo}}}.

\end{abstract}

\ccsdesc[500]{Computing methodologies~Motion capture}
\ccsdesc[500]{Computing methodologies~Neural networks}

\keywords{Character Animation, Motion Capture, Machine Learning}

\begin{teaserfigure}
  \includegraphics[width=\textwidth]{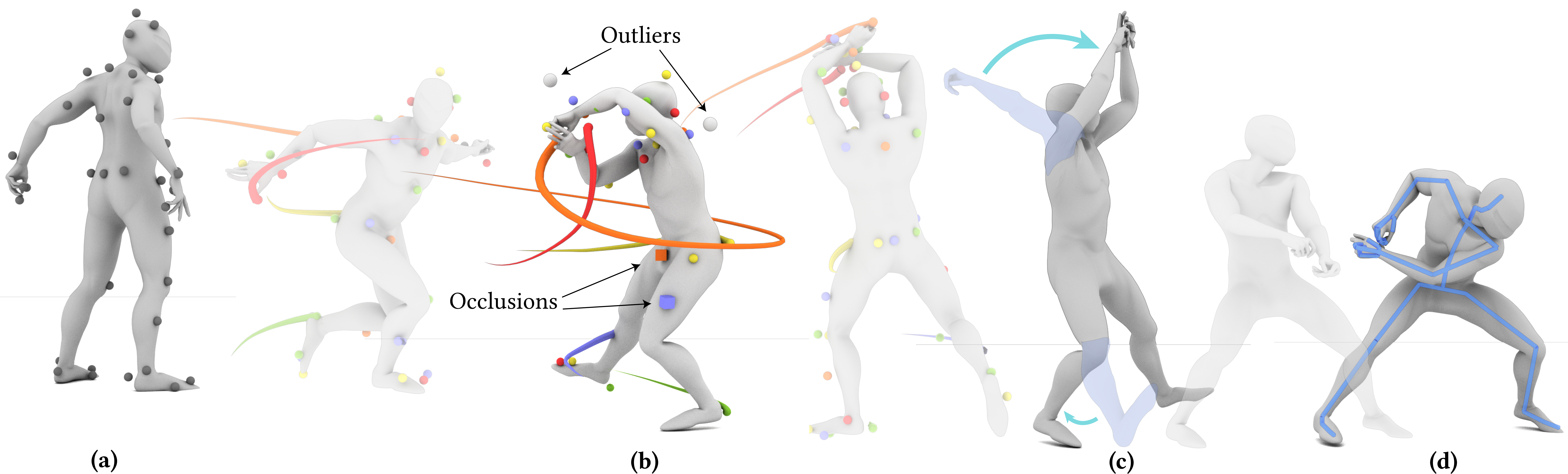}
  \caption{Given raw, unlabeled full-body motion capture point cloud (black balls) (a), RoMo generates tracklets to leverage temporal information for accurately labeling markers (colored balls) (b) and utilizes a hybrid inverse kinematics-based method (c) to solve body motions (d). RoMo demonstrates robustness against occlusion (cubes) and outliers (grey balls) and accurately solves large body motions and fine multi-fingered hand movements. }
  \label{fig:teaser}
\end{teaserfigure}

\maketitle

\section{Introduction}

Human pose estimation is a crucial field in computer vision and graphics. Despite the promising outcomes of marker-free, image-based methods, marker-based motion capture (MoCap) systems are preferred for their precision and versatility in game and film industry~\cite{GRAB} to capture body and hand motions with high coordination. MoCap system operates by first capturing a series of 2D images from multiple perspectives of the subjects, with markers at key locations on their bodies. The system then determines the 3D positions of these markers and associates them with specific locations of body parts through a process known as marker \textbf{labeling}. Next, the system reconstructs underlying skeletal structures and body movements, which is also known as \textbf{solving}.

Despite the use of sophisticated and expensive motion capture equipment, motion capture data inevitably suffers from various issues, including mislabeling, occlusions, and positional inaccuracies. These noises necessitate extensive manual corrections to achieve high accuracy. Noises can be categorized into three main types:

\begin{itemize} 
\item \textbf{Mislabeling}: These noises include label swaps, where markers are incorrectly labeled, and outliers, which occur when invalid markers are not properly excluded. Hand markers are susceptible to mislabeling due to the complexity of hand motions and their proximity.
\item \textbf{Occlusion}: This type of noise occurs when markers become invisible to the MoCap system, typically due to actors' self-occlusion or markers entering cameras' dead zones. 
\item \textbf{Positional inaccuracy}: These result from markers being recorded at positions that do not accurately reflect their true locations. Positional inaccuracies can occur when correcting occlusions or due to tracking system errors.
\end{itemize}

A multitude of data-driven approaches have been developed to address these noises, focusing on cleaning the raw marker data and solving motions. Some works~\cite{soma,ghorbani2019auto} concentrate on the labeling stage, but are limited to body markers alone. When these techniques are applied to whole-body marker data, the body and hand markers are treated equally, resulting in lower accuracy for hand markers due to distinct data distributions between body and hand markers, with the former having larger mutual distances and lower occlusion or mislabeling probabilities. Furthermore, these methods label the point cloud using information of a single frame, neglecting valuable temporal information. This omission reduces labeling accuracy, especially in the presence of occlusions. Another branch of research~\cite{holden18,MoCapSolver,pan23localmocap} aims to clean the noises in already labeled markers during the solving stage. These approaches often rely on the presumed accuracy of initial labeling or employ weak assumptions, such as acceleration thresholds~\cite{pan23localmocap} and marker distances~\cite{holden18,MoCapSolver}, to address mislabels. Consequently, they are susceptible to labeling errors. Moreover, all three types of noises can lead to inaccuracies in solving. These methods attempt to estimate the relative joint rotations of the entire body simultaneously, ignoring error accumulation along the kinematic chain of the body, where minor errors in parent joints can significantly amplify inaccuracies in child joints' positioning.

To develop a MoCap labeling and solving framework that is robust against the aforementioned noises, particularly mislabeling and positional inaccuracies, three key challenges must be addressed.
(1) The ability to deal with the distinct data distributions of body and hand markers. Even with deeper or wider neural networks, it is difficult for a single network to handle the full-body MoCap data containing body and hand markers.
(2) The exploitation of temporal continuity in marker data. A simple approach is to use the marker positions to create tracklets. Yet, this strategy overlooks the inter-marker relationships within the frame, leading to inaccuracies during rapid movements, such as rapid scrolling and fast hand waving. 
(3) The prevention of error accumulation along kinematic chain. Prior methods focus on the accuracy of joints' local rotations without considering the overall positional accuracy. It is crucial to devise a strategy that maintains awareness of the global positions of joints to prevent error accumulation.

To this end, we introduce RoMo, a data-driven approach for robustly labeling and solving full-body unlabeled optical MoCap data, mainly addressing two types of noises: mislabeling and positional inaccuracy. During the labeling stage, a divide-and-conquer strategy is employed to decompose the complex full-body labeling challenge into manageable subtasks: alignment, full-body marker segmentation and part-specific labeling. This approach mitigates training difficulties and enhances network performance. To make full use of temporal continuity of markers, tracklets are generated using deep features extracted by the network. Tracklet construction is modelled as a K-partite graph-based clustering algorithm, treating markers as nodes and edges are determined by feature similarities and positional differences. Incorporating feature similarities extends the receptive field of markers from nearest neighbors to all markers in the point cloud, thereby improving labeling precision. In the motion solving stage, RoMo uses joint positions as intermediate representations, with inverse kinematics adjusting the template skeleton to match the estimated joint positions. The utilization of global joint positions prevents error accumulation along the kinematic tree and enables accurate reconstruction of motions corresponding to input markers, especially the positions of ending joints. 


Overall, our paper makes the following contributions:

\begin{itemize}
    \item A robust full-body MoCap labeling framework that employs divide-and-conquer strategy and integrates spatial-temporal information to precisely label body and hand markers.
    \item An innovative method for generating marker tracklets by solving a K-partite graph-based clustering algorithm that utilizes both markers' positions and deep features.
    \item A hybrid inverse kinematics MoCap solver that circumvents error accumulation along the body kinematic chain.
\end{itemize}

\section{Related Work}

\begin{figure*}[htbp]
    \centering
    \includegraphics[width=\linewidth]{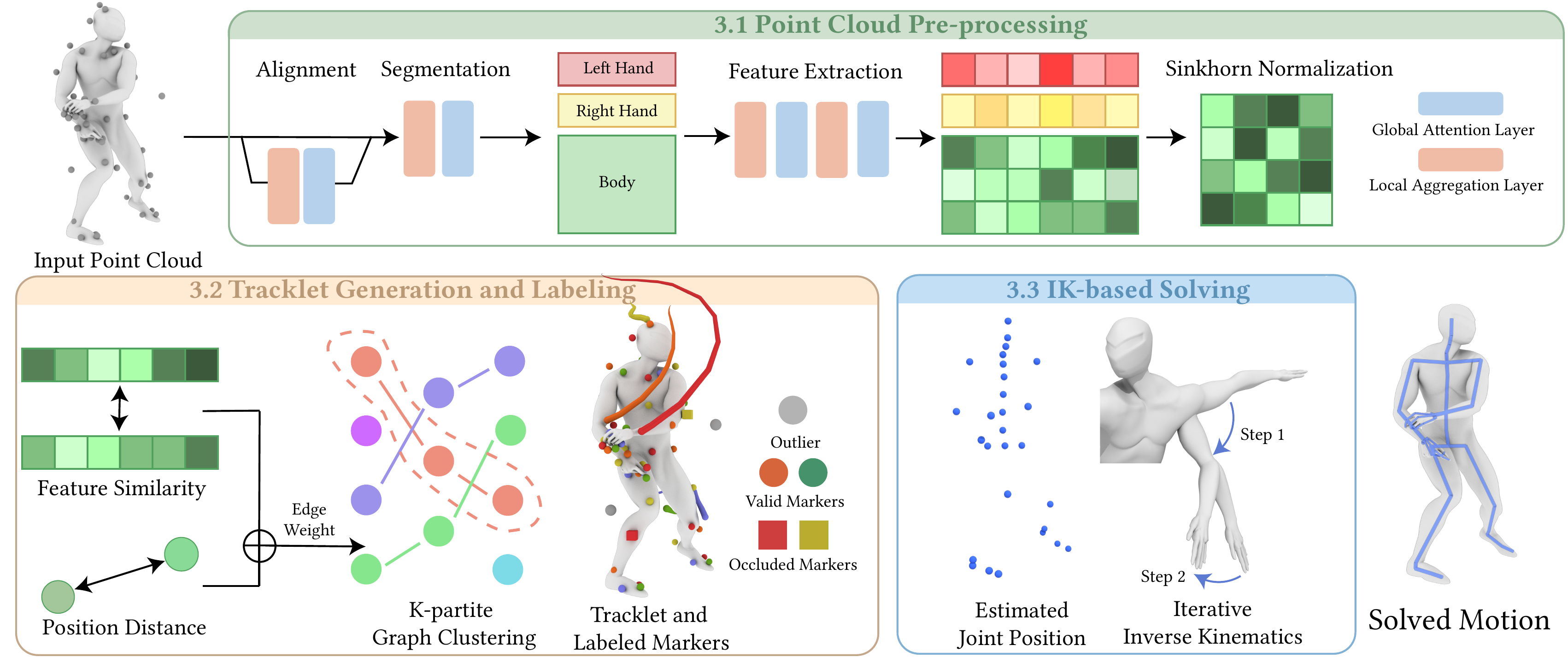}
    \caption{RoMo's pipeline consists of three modules. \textbf{Top}: In the pre-processing stage, RoMo accepts 3D sparse unordered MoCap point clouds with varying point numbers. It then conducts point cloud alignment to eliminate the global transformations and segmentation to partition the point cloud into body and hand point clouds. Subsequently, it employs a network consisting of alternating global self-attention and local aggregation layers to extract markers' features. \textbf{Bottom Left}: In the tracklet construction stage, RoMo addresses a K-partite graph-based clustering problem to create tracklets, and assigns markers within the same tracklet to a same label. \textbf{Bottom Right}: RoMo utilizes a hybrid inverse kinematics-based method to solve the motion, which iteratively adjust the joints of template skeleton along the kinematic tree to match the estimated joint positions.}
    \label{fig:pipeline}
\end{figure*}

Motion capture plays a crucial role in film and game industries, imbuing characters with vitality and distinctive personalities. Accurately capturing human motions is also essential across various motion-related research domains, including action recognition~\cite{hua2023part,yan2018spatial}, action prediction~\cite{cao2020long,cui21}, motion synthesis~\cite{tevet2022human,chen2023executing}, and image-based pose estimation~\cite{varol17_surreal,munea2020progress}. However, optical motion capture data inevitably suffers from mislabeling, occlusion, and positional inaccuracy, necessitating extensive manual corrections. Both the industry and research community have devoted significant efforts toward tackling these noises.

Within the industry, there are multiple mature commercial solutions available for real-time labeling and motion solving, such as Vicon Shogun~\cite{Vicon} and OptiTrack~\cite{OptiTrack}. However, these solutions are tailored to specific marker layouts and require careful calibrations for each new actor to accommodate diverse body shapes. Consequently, they do not offer a universally applicable solution for diverse scenarios requiring customized marker layouts, which are prevalent in numerous motion capture applications.

Numerous hand-crafted prior-based methods have been proposed in the research community ~\cite{baumann11,tautges11,liu2006estimation,aristidou2018self,wang2016,xiao2015sparse,tits2018robust} to clean MoCap data. Although these methods produce satisfying results on data with specific noise patterns via careful hand-tuning, they suffer from poor generalization ability on the real world data with more complex noises. Recent advances in research community have spotlighted the utilization of neural networks. Existing neural-based strategies primarily aim at solving motions from motion capture data~\cite{holden18,MoCapSolver,perepichka2019robust,pavllo2018real,pan23localmocap}, with a strong reliance on pre-labeled datasets. These methods often resort to some fragile heuristics to eliminate ghost points and struggle to address mislabeling caused by label swaps. Moreover, prior techniques attempt to solve the whole body joint's local rotations in a single pass, which frequently leads to cumulative errors along the body kinematic chain. In contrast, RoMo innovates by adopting global joint points as an intermediate representation and utilizing a hybrid inverse kinematics-based approach. This novel strategy ensures highly accurate solving results, markedly improving position accuracy for the terminal joints.

Few methods aim at labeling optical motion capture data using neural networks~\cite{soma,ghorbani2019auto}. Notably, Ghorbani \textit{et al.}~\cite{ghorbani2019auto} introduced a novel approach using a simple feed-forward residual network that employs permutation learning for body marker labeling. However, their method exhibits limitation in handling outliers. The introduction of SOMA~\cite{soma} marked a significant leap forward by leveraging an attention-based network architecture alongside an optimal transport layer, enhancing the accuracy of body marker labeling while efficiently managing outliers. Unlike previous methods that strive for a one-fits-all solution, RoMo employs a divide-and-conquer strategy, significantly enhancing labeling accuracy on full-body MoCap data. Furthermore, RoMo innovates by constructing tracklets through solving a K-partite problem based on proximities in marker positions and similarities in markers' deep features, thus efficiently utilizing the temporal continuity. Additionally, these methods primarily focus on global attention, overlooking the critical local interactions among adjacent markers. To bridge this gap, RoMo integrates stacked local aggregation~\cite{wang2019dynamic} and global attention layers~\cite{vaswani2017attention} to adeptly capture both local and global point features. 

\CHANGE{Tracklet is extensively studied in the field of Multiple Object Tracking (MOT) \cite{dai19,shen2018multiobject,sheng2018iterative,zhang2020long}, and RoMo generates tracklets to exploit temporal continuity in motion. The tracklets are short trajectories with high confidence, which reduces false positive detections. The key issue is correctly identifying objects across multiple frames. Many algorithms are proposed for determining the best association, including min-cost flow~\cite{ben14}, conditional random field~\cite{milan2013detection,choi2015near,xiang2020end}, and multiple hypothesis tracking \cite{kim2015multiple}. RoMo is the first method to use k-partite graph clustering in MoCap data processing, and it produces cutting-edge quantitative and qualitative results. Unlike previous works that process objects in videos, RoMo employs K-partite clustering on point cloud sequences with varying cardinalities.}

\section{Method}

RoMo takes as input a sequence of unlabeled motion capture point clouds, denoted as $\{P^1,...,P^T\}$, where each $P^i$ represents a sparse and unordered point cloud at time $i$, containing $n^i$ points. The cardinalities of these clouds $n^i$ changes over time due to marker occlusion or ghost points. Initially, RoMo assigns each point to a corresponding marker label $\{L_1, ..., L_N, null\}$, with $L_i$ signifying a valid label and $null$ indicating outliers. The labeled MoCap data is then structured as $M\in \mathbb{R}^{T\times N\times 3}$, representing the ordered positions of markers, alongside $O\in \{0, 1\}^{T\times N}$, which marks the visibility of markers. Subsequently, RoMo solves the motions, $R\in \mathbb{R}^{T\times (3 + K \times 9)}$, incorporating both the global translation of the body and the local rotations of each joint, as well as the underlying template skeleton $S\in \mathbb{R}^{K \times 3}$, representing the joints' offsets relative to their parent joints. An overview of RoMo's pipeline is presented in Fig. \ref{fig:pipeline}. In the following part of this section, we will first explain the labeling process, and then describe the solving framework.

Initially, marker features are extracted using a neural network comprised of alternating global self-attention~\cite{vaswani2017attention} and local aggregation~\cite{wang2019dynamic} layers to capture both global and local features $f$ (top of Fig. \ref{fig:pipeline}). To simplify training, before feature extraction, RoMo removes the global orientation of markers and segments the point cloud into three parts: body, left hand, and right hand. The alignment and segmentation employ networks that share a similar structure with the feature extraction network. After the feature extraction, RoMo performs Sinkhorn normalization~\cite{adams2011ranking} to transform the features $f$ to labeling confidences $c$ which satisfies a relaxed one-to-one correspondence.

To harness the temporal information, tracklets are constructed (bottom left of Fig. \ref{fig:pipeline}), and markers within the same tracklet are assigned identical labels. A tracklet depicts the motion of a single marker through 3D trajectories over a short period of time. The construction of tracklets is formulated as a K-partite graph-based clustering problem~\cite{zhang2020long}. Initially, a graph $\mathcal{G}$ is created, treating markers as nodes and forming edges based on similarities between markers, which include both positional and feature similarities. Subsequently, a clustering for $\mathcal{G}$ is sought using greedy algorithm. For the label association, RoMo calculates the tracklet's confidence using the L-q norm~\cite{ghorbani2019auto} of markers' confidence, and selects label with the highest confidence.

In the solving stage, motions are solved frame-by-frame (bottom right of Fig. \ref{fig:pipeline}). The joint position, denoted as $J\in \mathbb{R}^{T\times K \times 3}$, serves as an intermediate representation, with inverse kinematics applied to estimate joint rotations. The rotations are decomposed into twist and swing components, $R=R_{sw} R_{tw}$, where the swing component's axis is perpendicular to the bone direction, and the twist component's axis is parallel with the bone direction. Utilizing the labeled markers $M^i$ at frame $i$, RoMo estimates the joint positions $J^i \in \mathbb{R}^{K \times 3}$. This allows the calculation of a closed-form solution for the swing rotation component $R^i_{sw}$ by applying Rodrigues formula~\cite{askey20051839} on the relative joint positions with child joints. Additionally, RoMo estimates the twist angle $R^i_{tw}$, yielding the complete joint rotations. This process is iteratively performed along the kinematic chain to sequentially estimate joint rotations. The solving network also estimates the joint offsets of each frame $S^i$. To maintain bone length consistency throughout the motion sequence, the overall joint offset is determined by averaging estimated offsets across the motion sequence: $S_{i}={\rm mean}(S^1_{i},...,S^{T}_{i})$.

\subsection{Point Cloud Pre-processing}

Unlike previous methods~\cite{soma,ghorbani2019auto} that preserve body orientations, RoMo begins by normalizing the global body orientation, which aligns point clouds to accelerate network convergence and enhance the accuracy. An orthogonal transformation matrix is estimated through a smaller network and then directly applied to the input point cloud. This auxiliary network mirrors the architecture of the feature extraction network, employing similar components. The network is trained based on the rotation of the root joint, with regularization applied to ensure the output matrix $A$ closely approximates an orthogonal matrix:
\begin{equation}
    \begin{split}
        \mathcal{L}_{align}=||A^i-\hat{R}^i_{root}||_2+\lambda_{reg}||I-A^i{A^i}^T||_2.
    \end{split}
\label{eq:loss_align}
\end{equation}

Subsequently, the point cloud is segmented. Given that body and hand markers exhibit distinct data distributions, employing a unified network for feature extraction from the entire body's point cloud is overly complex. To simplify the learning process, a divide-and-conquer strategy is implemented, wherein the whole-body point cloud is first divided into body,left hand and right hand parts, followed by feature extraction. This segmentation process utilizes a neural network with a structure identical to that of the alignment network, and it is trained using cross-entropy loss.

Following the transformer-based point cloud processing networks~\cite{lu2022transformers}, RoMo's network architecture incorporates two types of layers: self-attention layers and local aggregation layers. Self-attention layers initiate by projecting the input features into key, query, and value vectors, and then employ the first two to calculate marker attention. The resultant layer output is derived from the product of attention weights and the value vector. Relying solely on the self-attention layers~\cite{soma} overlooks local spatial information, which is critical for point cloud processing. To address this, RoMo integrates local aggregation layers~\cite{wang2019dynamic} to capture the neighboring details. This process involves identifying the k-nearest neighbors within the feature space of points, and utilizing feature differences to aggregate relative information regarding neighboring markers. Such data are subsequently aggregated and processed through a multi-layer perceptron (MLP) to produce the local aggregation layer's output.

The confidence of labels to markers is designed to satisfy a relaxed one-to-one correspondence: each point can be mapped to at most one label and vise versa, with the exception of the $null$ label, which can be assigned to multiple points. To enforce this constraint, RoMo employs log-domain optimal transport~\cite{peyré2020computational, sarlin2020superglue}, utilizing the Sinkhorn-Knopp algorithm iteratively for optimization~\cite{adams2011ranking}. Following ~\cite{soma}, RoMo constructs the initial confidence matrix $C^i_{init} \in [0, 1]^{n^i\times N}$, using the extracted marker features $f$, and modifies it by adding an extra row and column to account for unmatched points and labels, resulting in the augmented confidence matrix ${C^i}_{aug} \in [0, 1]^{(n^i+1)\times (N+1)}$. Sinkhorn normalization is then iteratively applied to alternatively normalize the rows and columns, ensuring they sum to 1. After $k$ iterations, we obtain the final normalized confidence matrix $C^i$.

The feature extraction network's training incorporates a weighted negative likelihood loss: $\mathcal{L}_{label}=-\frac{1}{\sum \hat{C^i}}\sum W \hat{C^i} \log(C^i)$, with $\hat{C^i}$ as the ground truth confidence matrix and $W$ serving as a weight matrix to mitigate the impact of the $null$ label.

\subsection{Tracklet Generation and Labeling}

The previous section focused solely on information within a single frame, overlooking the temporal continuity of marker motions. To harness temporal information in sequential MoCap data, RoMo generates tracklets by leveraging the marker positions and marker features extracted by neural network. While some commercial software, such as Vicon~\cite{Vicon}, automatically generates tracklets, these are not universally available, notably in certain archival MoCap datasets like GRAB~\cite{GRAB} or limited hardware. Additionally, these methods do not utilize deep features of markers, limiting the precision of tracklet construction.

Drawing inspiration from ~\cite{zhang2020long}, RoMo treats tracklet construction as a K-partite graph-based clustering algorithm. After construction, RoMo assigns the markers in the same tracklet with the same label, determined by the L-q norm of confidence of the labels computed in the preceding subsection.

Conceptually, each marker is considered a node within the graph, and edges are established between nodes that exhibit high similarity across different frames. Specifically, for a MoCap point cloud sequence, the constructed graph $\mathcal{G}=\{\mathcal{V}, \mathcal{E}, \mathcal{W}\}$ consists of the following three components:

\begin{itemize}
    \item Node set $\mathcal{V}$ represents the markers in the point cloud, where $v^i_m$ denotes the m-th marker in the point cloud $P^i$.
    \item Edge set $\mathcal{E}$ comprises edges $e_{mn}^{ij}$ indicates temporal connections, which connect any two nodes ($v^i_m, v^{j}_n$) that represent markers from different frame,  
    \item Weight set $\mathcal{W}$: the weight $w_{mn}^{ij}$ associated with edge $e^{ij}_{mn}$ quantifies the similarity between markers $v^i_m$ and $v^{j}_n$, incorporating both spatial proximity and feature resemblance to measure marker similarity.
\end{itemize}

\begin{equation}
    \begin{split}
        w^{ij}_{mn} & =w^{ij}_{pos,mn}+\lambda_{fet} w^{ij}_{fet,mn}, \\
        w^{ij}_{pos,mn} & =||p^i_m,p^j_n||_2, \\
        w^{ij}_{fet,mn} & =1-{\rm cosine\_similarity}(f^i_m,f^j_n),
    \end{split}
\label{eq:graph_weight}
\end{equation}
where $w_{pos,mn}$ represents the Euclidean distance between marker positions, while $w_{fet,mn}$ quantifies the cosine distance between the deep features of the corresponding nodes, with $\lambda_{fet}$ serving as a predefined weighting factor of feature similarity.

To identify a clustering solution for the graph $\mathcal{G}$, RoMo seeks to minimize the total sum of all weights, as outlined below:

\begin{equation}
    \begin{split}
        \arg \min_{e^{ij}_{mn}} & \sum_{m,n} e^{ij}_{mn}w^{ij}_{mn}; \\
        s.t.\ \sum_{v^j_n\in \mathcal{V}^i, i\neq j} e^{ij}_{mn} \leq 1 &;\quad \sum_{v^j_n\in \mathcal{V}^i, i = j} e^{ij}_{mn} = 0; \\
        e^{ij}_{mn} = 0, w^{ij}_{pos,mn}>th_{pos} &;\quad e^{ji}_{mn} = 0, w^{ij}_{fet,mn}>th_{fet}; \\
        e^{ij}_{mn} \in \{0, 1\} &; \quad e^{ij}_{mn}=e^{ji}_{nm};
    \end{split}
\label{eq:K-partite}
\end{equation}
where $\mathcal{V}^i$ refers to the node set of the i-th frame, while $th_{pos}$ and $th_{fet}$ are thresholds for position and feature differences, respectively. The first condition ensures that any given node $ v_m^i $ connects to at most one node in the node set of a different frame $ \mathcal{V}^j $. The edge value $e^{ij}_{mn}$ is restricted to either 0 or 1, resulting in integer solutions for this optimization problem. To expedite computation, a greedy algorithm is employed. In the resulting solution, a collection of cliques is obtained, which are considered as tracklets.

For the association of tracklets with marker labels, a straightforward approach is to assign the most frequently occurring label from single-frame labeling, yet this method overlooks scenarios involving labels that are prevalent but have low confidence. To circumvent this issue, RoMo's tracklet labeling method calculates tracklet confidence from markers' confidence. Specifically, for a tracklet $Tr=\{p^i_l, ...,p^j_m, ..., p^k_n\}$, the confidence of it being assigned to marker label $L_i$ is calculated as $C_{L_i}=(\sum_{p^i_l \in Tr}|c_{p^i_l,L_j}|^q)^{\frac{1}{q}}$, representing the $L-q$ norm of the confidence for label $L_i$ of markers within $Tr$. For $q=0$, this score mimics a voting mechanism, whereas for $q=1$, it serves as the sum of confidence of label $L_i$.

\subsection{Hybrid Inverse Kinematics-based Solving}

\begin{figure}[htbp]
    \centering
    \includegraphics[width=\linewidth]{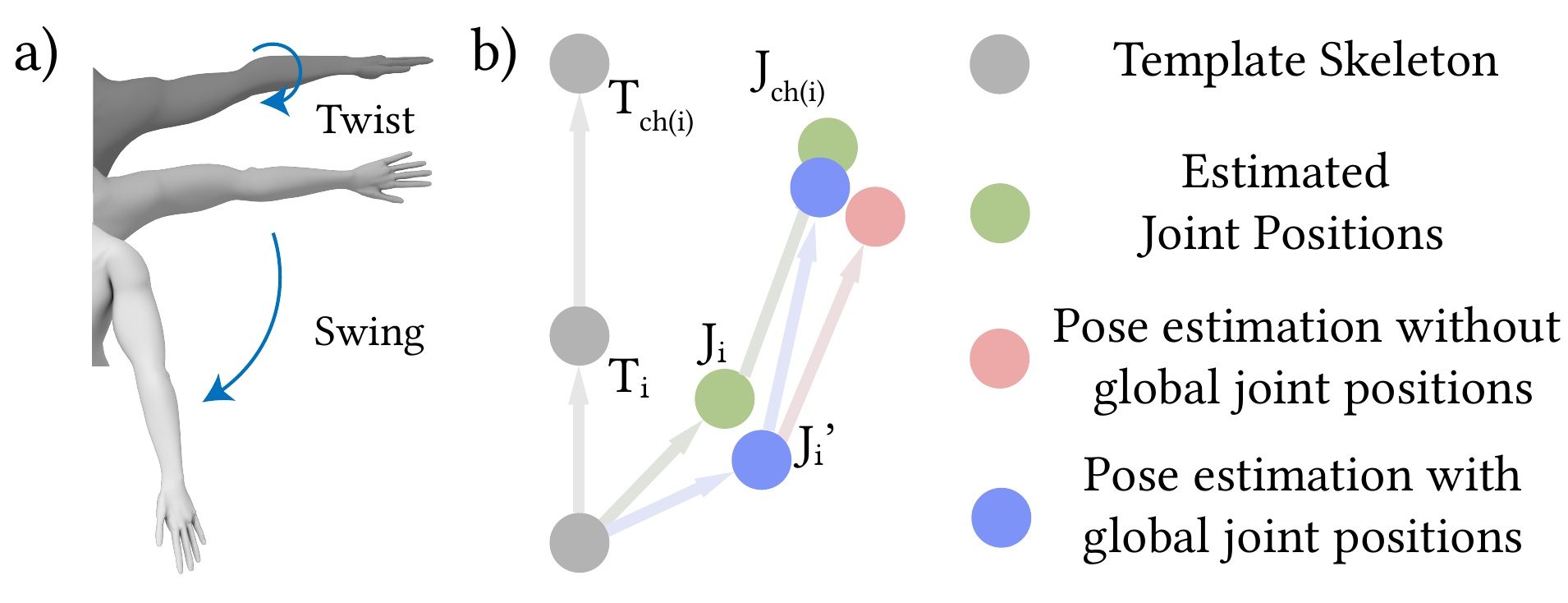}
    \caption{a) An illustration of rotation decomposition. b) RoMo utilizes global joint positions and avoids error accumulation along kinematic chain.}
    \label{fig:hybrik}
\end{figure}

Directly estimating local joint rotations in one pass may cause error accumulation along the kinematic chain,  where minor rotational inaccuracies in parent joints can lead to significant positional discrepancies in child joints, as shown in the right of Fig.~\ref{fig:hybrik}. This process is further challenged by high non-linearity, as rotations belong to the 3D rotation group $R\in {\rm SO}(3)$ and marker positions belong to the 3D position group $M\in R^3$, which complicates network training due to its inherent difficulty~\cite{kolotouros2019learning,kanazawa2018end}. Inspired by~\cite{li2021hybrik}, RoMo leverages 3D keypoints as an intermediate representation and uses an inverse kinematics-based solver. This approach simplifies network training and mitigates error accumulation. \CHANGE{Compared to \cite{li2021hybrik}, RoMo employs a different feature extractor for the 
 task of optical motion capture solving.} Next, we dive into details of the solver, the times in notations are omitted for simplicity.


The key idea is to estimate 3D joint positions rather than the complete rotations. With the estimated 3D joint positions $J$, a solution for the swing rotation $R_{sw}$, whose axis is perpendicular to bone direction, is obtainable. The twist rotation $R_{tw}$, whose axis is parallel to the direction, however, requires the assistance of a neural network. An illustration of rotation decomposition is shown in the left of Fig. \ref{fig:hybrik}.  The full rotation is the combination of the swing and twist rotation $R=R_{sw}R_{tw}$. Although this method still relies on the neural network for rotation estimation, it significantly reduces learning complexity. Unlike the three degree of freedoms (DoF) associated with rotation, the twist angle is constrained to a single DoF. Additionally, due to the physical constraints of the human body, the twist angle exhibits a limited range of variation~\cite{li2021hybrik}. Only the joints on limbs exhibit a wide range of twist angles, constituting a small proportion of all the joints. We present a detailed analysis of joints' twist angle's distribution in the supplementary material.

The hybrid inverse kinematics process is performed iteratively along the kinematic tree. \CHANGE{
In our experiments, we discovered that using only the network on the point cloud to estimate the root orientation can lead to an error of about 2 degrees. As a result, we use Singular Value Decomposition (SVD) along with the estimated joint positions to improve the root orientation solving accuracy.
} 
For the estimation of rotations of subsequent children joints, we calculate the swing rotation $R_{sw}$ from the estimated joint positions. A straightforward approach is to employ the relative positions derived from the estimated positions of child joints:
\begin{equation}
    \begin{split}
        \vec{j}_i = J_{ch(i)}-J_i &, \quad \vec{t}_i = T_{ch(i)}-T_i, \\
        {\rm cos} \alpha_i  = \frac{\vec{j}_i \cdot \vec{t}_i}{||\vec{j}_i ||\ || \vec{t}_i||} &, \quad
        {\rm sin} \alpha_i  =\frac{||\vec{j}_i \times \vec{t}_i||}{||\vec{j}_i ||\ || \vec{t}_i||}, \\
                \vec{n}_i  &= \frac{\vec{j}_i \times \vec{t}_i}{||\vec{j}_i \times \vec{t}_i||} ,\\
        R_{sw,i} = \textit{I} \ + \ & {\rm sin} \alpha_i [\vec{n}_i]_{\times} \ + \  (1- {\rm cos}\alpha_i) [\vec{n}_i]^2_{\times}, 
    \end{split}
\label{eq:rodrigues}
\end{equation}
where $\vec{j}_i, \vec{t}_i$ are the estimated and template relative joint positions, respectively. The rest is the Rodrigues formula, where $\textit{I}\ $ is the identity matrix and $[\vec{n}_i]_{\times}$ is the skew symmetric matrix of $\vec{n}_i$.

However, this approach does not mitigate error accumulation as it relies solely on the relative positions of child joints, disregarding errors introduced by parent joints. To account for such discrepancies, RoMo employs forward kinematics on the template skeleton with the current estimated joint rotations to obtain the global joint positions $J'_i$, which can also reflect the error introduced by estimated parent joints' offsets and rotations. The relative joint position is then calculated as $\vec{j}'_i = J_{ch(i)} - J'_i$. We replace the first equation in Equation \ref{eq:rodrigues} using $\vec{j}'_i$. This process is illustrated in the right of Fig. \ref{fig:hybrik}.

RoMo's motion solver uses the heterogeneous graph network outlined in~\cite{pan23localmocap} as its backbone. This network considers markers and joints as distinct nodes in a heterogeneous graph and conducts graph convolution operations to extract their local and global features to solve motions. The motion solving network estimates joint positions $P$, twist rotations $R_{tw}$, and bone lengths $S$. For continuity, the twist rotation is represented by a two-dimensional vector $[\cos(R_{tw}), \sin(R_{tw})]$, derived from the trigonometric functions of the angle. Direct computation of bone lengths from joint positions was explored but resulted in lower accuracy. The network's training is based on a set of loss functions specified below:
\begin{equation}
    \begin{split}
        \mathcal{L}_{solving}=\lambda_{pos}||J-\hat{J}||_2+\lambda_{tw}||R_{tw}-\hat{R}_{tw}||_2+\lambda_{skel}||S-\hat{S}||_2,
    \end{split}
\label{eq:loss_solving}
\end{equation}
where $\hat{J}, \hat{R}_{tw}, \hat{S}$ are ground truth joint positions, twist rotations and joint offsets, respectively. $\lambda_{pos},\lambda_{tw},\lambda_{skel}$ are weight factors.

\section{Experiments}

\definecolor{MyGray}{HTML}{D8D8D8}

\begin{table*}[htbp]

\scalebox{0.85}{
\begin{tabular}{cc|cc|cc|cc||c|cc|cc|cc|cc|cc}
            &      & \multicolumn{2}{c|}{\rotatebox{0}{\tabincell{c}{[Ghorbani et \\ al. 2019]}}} & \multicolumn{2}{c|}{\rotatebox{0}{\tabincell{c}{SOMA}}} & \multicolumn{2}{c||}{\rotatebox{0}{\tabincell{c}{RoMo}}} &       & \multicolumn{2}{c|}{\rotatebox{0}{\tabincell{c}{Vicon / \\ MoSh++}}} & \multicolumn{2}{c|}{\rotatebox{0}{\tabincell{c}{[Holden \\ 2018]}}} & \multicolumn{2}{c|}{\rotatebox{0}{\tabincell{c}{MoCap-Solver}}} & \multicolumn{2}{c|}{\rotatebox{0}{\tabincell{c}{LocalMoCap}}} & \multicolumn{2}{c}{\rotatebox{0}{\tabincell{c}{RoMo}}} \\
            \hline
\multirow{2}{*}{Production}  & F1 $\uparrow$  &    99.69          &   \cellcolor{MyGray}  92.28          &      99.85       &      \cellcolor{MyGray}  94.58    &    \textbf{99.94}         &   \cellcolor{MyGray}   \textbf{98.62}      & MPJRE $\downarrow$ &        4.21     &    \cellcolor{MyGray} 1.37         &   2.54           &   \cellcolor{MyGray}   0.80       &     1.89            &   \cellcolor{MyGray} 0.58           &       1.22         &    \cellcolor{MyGray}  0.45         &      \textbf{1.09}       &    \cellcolor{MyGray}  \textbf{0.42}      \\
            & Acc. $\uparrow$ &     99.59          &   \cellcolor{MyGray}  93.02         &      99.87       &     \cellcolor{MyGray}  96.21     & \textbf{99.96}            & \cellcolor{MyGray} \textbf{98.87}           & MPJPE  $\downarrow$  &     1.75        &   \cellcolor{MyGray}  0.49        &     1.02         &    \cellcolor{MyGray}   0.27      &   0.89              &    \cellcolor{MyGray}   0.20         &   0.61            & \cellcolor{MyGray}  0.15            &  \textbf{0.43}           &  \cellcolor{MyGray}   \textbf{0.14}       \\
            \hline
\multirow{2}{*}{Front Waist} & F1 $\uparrow$  &       97.37        & \cellcolor{MyGray} 87.29            &    98.26         & \cellcolor{MyGray} 90.24          &  \textbf{99.46}           &  \cellcolor{MyGray}  \textbf{97.86}        & MPJRE $\downarrow$ &      3.93       &  \cellcolor{MyGray} 1.44          &     2.79         &     \cellcolor{MyGray}  1.02      &        2.25         &    \cellcolor{MyGray}   0.70         &       1.75         &    \cellcolor{MyGray}  \textbf{0.60}         &     \textbf{1.55}        &   \cellcolor{MyGray}  0.62       \\
            & Acc. $\uparrow$ &       98.59        &    \cellcolor{MyGray} 92.18         &        99.28     &     \cellcolor{MyGray}  93.87     &       \textbf{99.68}      &  \cellcolor{MyGray}  \textbf{98.23}        & MPJPE $\downarrow$   &      1.66       &     \cellcolor{MyGray} 0.30       &       1.11      &   \cellcolor{MyGray} 0.32         &   0.99              &     \cellcolor{MyGray}  0.25         &        0.82        &     \cellcolor{MyGray} 0.21         &    \textbf{0.69}         &  \cellcolor{MyGray} \textbf{0.21}         \\
            \hline    
\multirow{2}{*}{GRAB} & F1  $\uparrow$ &  99.65             &    \cellcolor{MyGray} 94.48         &        \textbf{99.70}     &  \cellcolor{MyGray}   96.92       & 99.65            &  \cellcolor{MyGray} \textbf{98.47}         & MPJRE $\downarrow$ &       3.32      &     \cellcolor{MyGray}  6.37      &  2.69            &  \cellcolor{MyGray} 2.21         &        2.15         &   \cellcolor{MyGray} 1.72            &       1.83         &    \cellcolor{MyGray} \textbf{1.59}          &      \textbf{1.70}       &   \cellcolor{MyGray} 1.61        \\
& Acc. $\uparrow$ &      \textbf{99.72}         &      \cellcolor{MyGray}  95.57      &          99.68   &      \cellcolor{MyGray}  97.27    &    99.69         &   \cellcolor{MyGray} \textbf{98.69}         & MPJPE $\downarrow$   &     1.79        &       \cellcolor{MyGray} 0.79     &     1.21         &      \cellcolor{MyGray}  0.27     &      1.13          &    \cellcolor{MyGray}  0.21          &     0.98           &           \cellcolor{MyGray} \textbf{0.18}   &    \textbf{0.83}         &   \cellcolor{MyGray} 0.19        \\
\end{tabular}
}

\caption{Comparison with other methods on labeling (left) and solving (right). Cells with white background display metrics for body, those with gray background represent metrics for \colorbox {MyGray}{hand.} The F1 and accuracy are multiplied by 100. The units of MPJRE and MPJPE are in degree and centimeter, respectively.}
\label{table:comparison}
\end{table*}

We evaluate RoMo quantitatively on several full-body MoCap datasets. The Production and Front-waist dataset are captured from a game studio and cleaned by hand, with different marker layouts. We take the raw markers as input and clean markers as ground truth. The quality of these datasets is exceptionally high, as they are captured in a professional setting with a large number of cameras. The occlusion probabilities for body and hand markers are 0.5\% and 6\%, respectively. The GRAB dataset~\cite{GRAB} captures people interacting with everyday objects. We simulate noises by adding occlusions, where different markers on different body parts have diverse occlusion probabilities, which are propositional to the real data. To emulate data captured under in limited conditions, we set the overall occlusion rate at 15\%. The supplementary material contains additional information about our dataset, network architectures, hyper-parameters and implementation details.

\subsection{Point Cloud Labeling}

\begin{table}[htbp]
\scalebox{0.92}{
\begin{tabular}{c|cc|cc}
     & \multicolumn{2}{c|}{F1} & \multicolumn{2}{c}{Accuracy} \\
     \hline
Base &     \multicolumn{2}{c|}{\cellcolor{MyGray}  98.62}          &    \multicolumn{2}{c}{\cellcolor{MyGray} 98.87}              \\
 - Local aggregation layers     &  \cellcolor{MyGray}   98.59        &    \cellcolor{MyGray}     -0.03   &    \cellcolor{MyGray}     98.32      &     \cellcolor{MyGray}   -0.55    \\
  - Point cloud segmentation     &   \cellcolor{MyGray}  95.20        &    \cellcolor{MyGray}   \textbf{-4.74}     &      \cellcolor{MyGray}  95.31      &     \cellcolor{MyGray}   \textbf{-3.56}     \\
  - Global transformation removal    &   \cellcolor{MyGray}  98.10        &   \cellcolor{MyGray}   -0.52   &   \cellcolor{MyGray}  98.23         &   \cellcolor{MyGray}    -0.64      \\
 - Tracklet-based labeling    &   \cellcolor{MyGray}  96.21        &    \cellcolor{MyGray}    \textbf{-2.41}    &  \cellcolor{MyGray}  96.79          &  \cellcolor{MyGray}    \textbf{-2.08}      \\
 - Feature similarity in edge weights    &   \cellcolor{MyGray}  98.01        &     \cellcolor{MyGray}   -0.61    &     \cellcolor{MyGray} 98.38       &    \cellcolor{MyGray}      -0.49       
\end{tabular}
}
\caption{Ablation study of RoMo's components on the Production dataset's hand markers. We take the full model as the baseline and remove one component at a time.}
\label{table:ablation_label}
\end{table}

For marker labeling, we compare RoMo with two neural-based approaches: \cite{ghorbani2019auto} and SOMA~\cite{soma}. We report two metrics: F1 score and accuracy in percentages. F1 score is the harmonic-average of the precision and recall score, and accuracy is the proportion of correctly predicted labels over all labels. Since the body and hand markers have distinct data distributions, we separately display their metrics to clearly show the methods' performance on diverse body parts.

\begin{figure}[htbp]
    \centering
    \includegraphics[width=\linewidth]{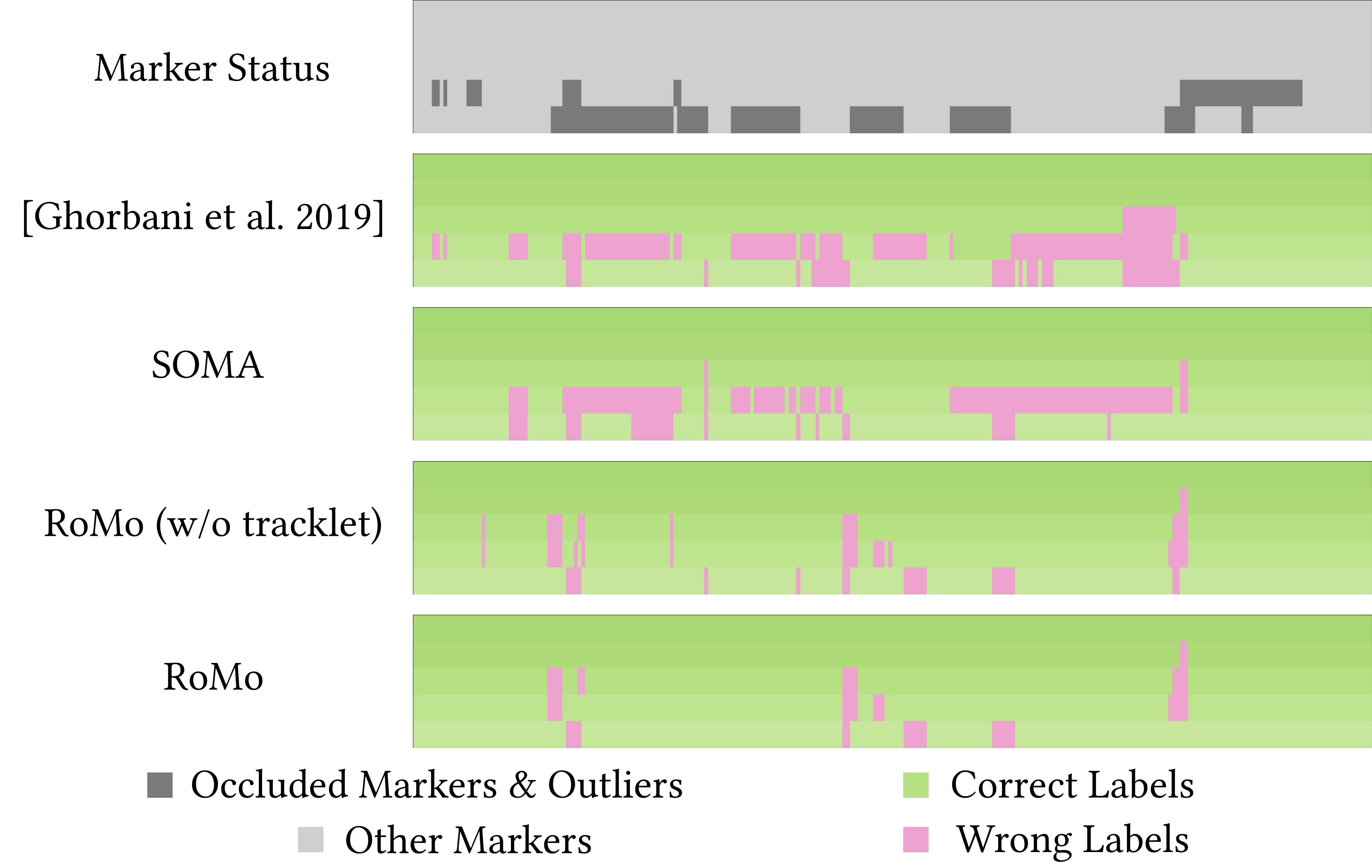}
    \caption{A qualitative comparison with other methods on hand labeling, where each column represents a timestamp in a MoCap sequence. In cases of outliers, we either randomly place them in positions corresponding to occluded markers or discard them if there isn't enough space.}
    \label{fig:label_qualitative}
\end{figure}

\begin{figure*}[htbp]
    \centering
    \includegraphics[width=\linewidth]{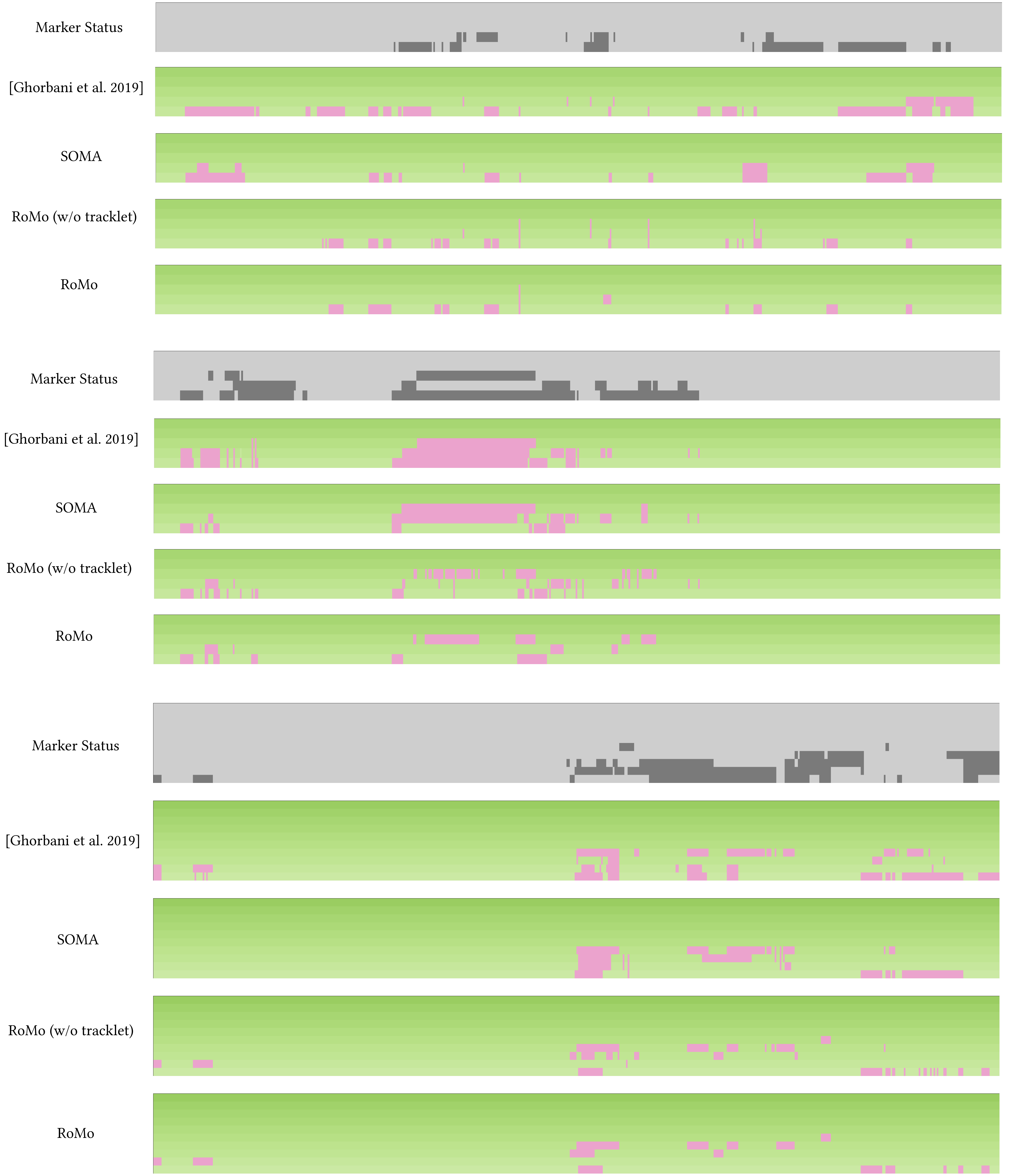}
    \caption{Additional qualitative comparison of labeling.}
    \label{fig:additional_labeling_results}
\end{figure*}

We present the quantitative result on the left of Table~\ref{table:comparison}. Thanks to the high data quality, the accuracy of body marker labeling in the Production and Front-waist datasets exceeds 98\%, with RoMo marginally outperforming previous methods. However, the accuracy for hand data in comparing methods drops dramatically to about 94\%, which can be attributed to two main factors. Firstly, the distinct data distributions between body and hand data make it challenging for a single network to effectively address both types. Secondly, hand data are have higher occlusion probabilities and closer marker proximities. Frame-by-frame labeling failed to handle such complex data. Our method maintains high labeling accuracy through a divide-and-conquer strategy and the use of a tracklet-based labeling schema. We further present a qualitative comparison in Fig.~\ref{fig:label_qualitative}. RoMo demonstrates robustness to occlusions and outliers and achieves the highest accuracy compared to other methods, particularly under complex occlusions. Additionally, the utilization of tracklets significantly boosts labeling performance. We present additional qualitative comparisons on Fig.~\ref{fig:additional_labeling_results}.

Table \ref{table:ablation_label} shows the effect of various components of RoMo on the validation split of the Production dataset. The point cloud segmentation and tracklet-based labeling play the most significant role in the overall performance of marker labeling. Other components marginally improves the performance of the model.

\subsection{Motion Solving}

For motion solving, we compare RoMo with three neural-based approaches ~\cite{holden18,pan23localmocap,MoCapSolver}. Before solving, we apply the state-of-the-art occlusion fixing algorithm ~\cite{pan23localmocap} to fill the missing markers, which first uses the Euclidean distance matrix optimization algorithm~\cite{zhou2020} and then employ a bidirectional long short time memory network (BiLSTM)~\cite{bilstm} to optimize the occluded markers' potisions. To ensure a fair comparison, the body and hand markers are aligned using the same strategy and trained using separate networks for neural approaches. Additionaly, we compare our method with two optimization-based methods: Vicon~\cite{Vicon} for Production and Front-waist dataset, and Mosh++~\cite{mosh} for the GRAB dataset. We use two quantitative metrics: mean per joint rotation error (MPJRE) and mean per joint rotation error (MPJPE). The former reflects the angle differences of the joints and latter represents the global joint position difference.

The quantitative results are displayed on the right in Table~\ref{table:comparison}. For body motion, RoMo's MPJRE and MPJPE are approximately 15\% and 25\% lower, respectively, than those achieved by the best comparison methods. RoMo outperforms other neural-based methods for two primary reasons. Firstly, it employs joint positions as intermediate representations, simplifying the learning process for neural networks. Secondly, RoMo leverages the inverse kinematics process to utilize the global joint positions, effectively avoiding the error accumulation along the kinematic tree. Other methods ~\cite{holden18,pan23localmocap,MoCapSolver} directly estimates the relative joint rotation of the whole body simultaneously. They are fragile to error accumulation, resulting in large discrepancy on ending joints. RoMo's solving accuracy of hand motions are comparable with the state-of-the-art methods, since the hand bone length of are relatively limited, and does not amplify parental joints' rotation errors.

\begin{figure}[htbp]
    \centering
    \includegraphics[width=\linewidth]{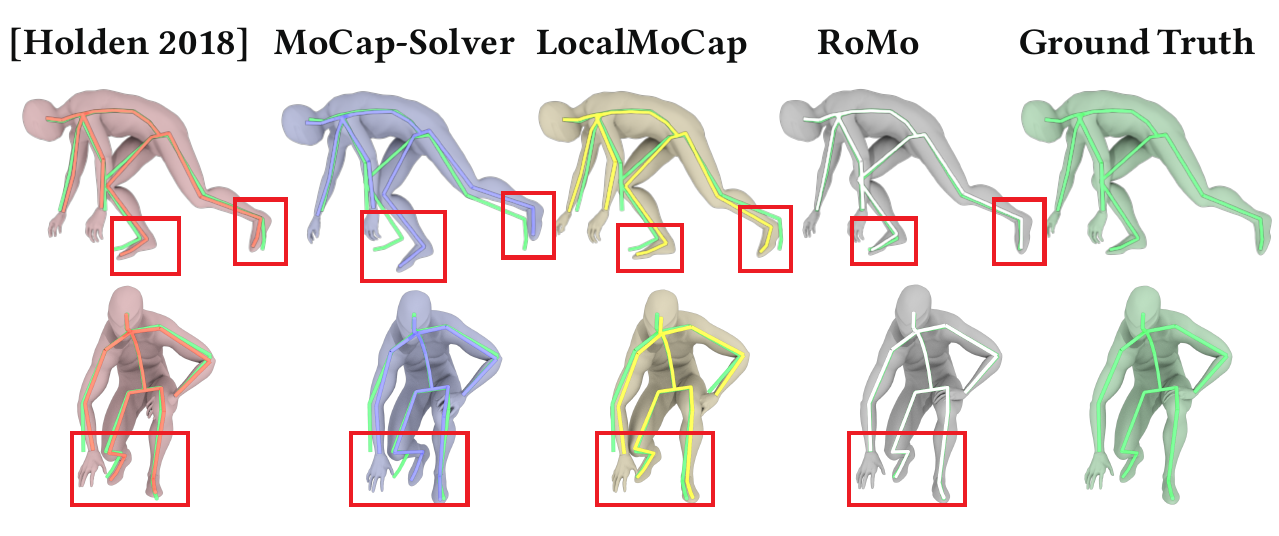}
    \caption{Qualitative comparisons of solved motions. To compare with the ground truth, we render the skeletons and overlay the ground truth’s skeleton onto those generated by solving methods. Positions with significant differences are indicated with red boxes.}
    \label{fig:comparison}
\end{figure}

\begin{figure*}[htbp]
    \centering
    \includegraphics[width=\linewidth]{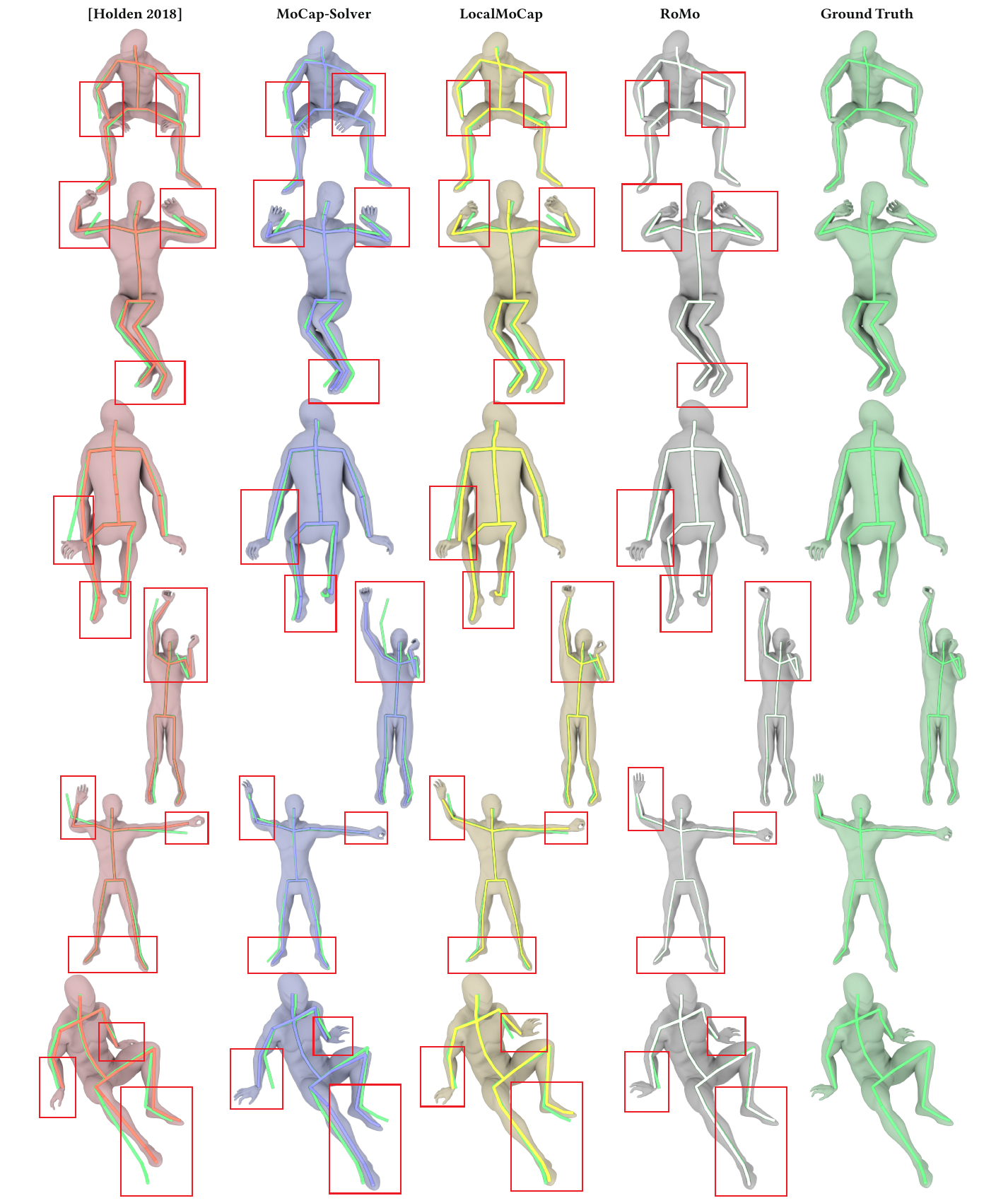}
    \caption{Additional qualitative comparisons of solved motions.}
    \label{fig:additional_comparison}
\end{figure*}

We present the qualitative results of body motion solving in Fig.~\ref{fig:comparison} and highlight RoMo's improvements. RoMo demonstrates better solving accuracy, especially on ending joints such as wrists and ankles. We present additional qualitative comparison on Fig.~\ref{fig:additional_comparison}.

To better demonstrate RoMo's robustness against positional errors, we introduce random jitters to the input marker positions and record the resulting joint position errors. We compare RoMo with a state-of-the-art solution method~\cite{pan23localmocap}. Additionally, we validated the use of global joint positions $J'$, with the results presented in Table~\ref{table:robustness}. Methods that rely solely on joints' relative rotations~\cite{pan23localmocap} or relative positions tend to suffer from errors accumulation along the kinematic chain, leading to increased errors at higher jitter intensities. In contrast, RoMo, by using global joint positions, exhibits greater robustness to such noise.

\begin{table}[htbp]
\scalebox{0.95}{
\begin{tabular}{c|cccc}
     & 0 & $\pm 0.2 \rm cm $& $\pm 0.5 \rm cm$ & $\pm 1.0 \rm cm$\\
     \hline
LocalMoCap~\cite{pan23localmocap}     &  0.61 &  0.72 &   0.98 &  1.62  \\
RoMo (w/o global joint position)     & 0.47  &  0.68 &   0.83 &   1.49 \\
RoMo & 0.43  &  0.62 & 0.76   & 1.25  
\end{tabular}
}
\caption{The MPJPE under various jitter intensities on the body data of Production dataset.}
\label{table:robustness}
\end{table}

\section{Conclusion, Limitations and Future Work}

We introduce RoMo, a framework designed for robust labeling and solving of optical motion capture data of full-body, addressing two primary types of noise: mislabeling and positional errors. RoMo utilizes innovative techniques such as a tracklet generation algorithm based on positional and feature similarities, coupled with an inverse kinematics-based motion capture solver to avoid error accumulation. Our evaluations on diverse and complex benchmarks demonstrate RoMo's superior performance compared to previous methods, providing precise labeling of body and hand markers and robustly solving full-body motions. Notably, RoMo achieves accuracy comparable to commercial systems but offers the advantages of being cost-free and highly flexible.

RoMo does have its limitations. Currently, it still splits the labeling and solving stages, since the input of labeling stage is unordered markers while the solving relies on labeled marker. Merging the two stages, that is, solving motions directly from unordered point cloud data, could diminish the reliance on heuristic assumptions and improve the labeling and solving network's generalization capabilities. Furthermore, integrating motion information into the labeling stage could potentially enhance label accuracy. Additionally, RoMo does not account for variations in marker layouts. Training directly on a layout superset encompassing all markers of diverse layouts~\cite{soma} substantially reduces labeling accuracy. Exploring solutions to accommodate marker layout variations within a single network without compromising performance presents a promising direction to augment RoMo's flexibility.

\begin{acks}

Xiaogang Jin was supported by Key R\&D Program of Zhejiang (No. 2023C01047) and the National Natural Science Foundation of China (Grant No. 62036010). We thank Guanglong Xu, Xianli Gu and other colleagues in Tencent Games Digital Content Technology Center for preparing the training data, discussing the result and rendering the video.

\end{acks}

\bibliographystyle{ACM-Reference-Format}
\bibliography{references}

\end{document}